%% file: main.tex
\documentclass[
]{ceurart}

\usepackage{listings}
\usepackage[numbers]{natbib}
\usepackage{placeins} 

\begin{document}

\copyrightyear{2026}
\copyrightclause{Copyright for this paper by its authors.
Use permitted under Creative Commons License Attribution 4.0
International (CC BY 4.0).}


\conference{{D}e{F}actify 4.0: Fourth workshop on Multimodal Fact-Checking and Hate Speech Detection, March 2025, Philadelphia, Pennsylvania, USA}

\title{Findings of the Counter Turing Test: AI-Generated Text Detection}

\tnotemark[1]
\tnotetext[1]{This document is based on the CEUR-WS template and incorporates topics inspired by the Defactify workshop series.}

\author[1]{Rajarshi Roy}[email=royrajarshi0123@gmail.com]
\author[5]{Gurpreet Singh}
\author[2]{Ashhar Aziz}
\author[3]{Shashwat Bajpai}
\author[4]{Nasrin Imanpour}
\author[6]{Shwetangshu Biswas}
\author[7]{Kapil Wanaskar}
\author[8]{Parth Patwa}
\author[9]{Subhankar Ghosh}
\author[10]{Shreyas Dixit}
\author[1]{Nilesh Ranjan Pal}
\author[4]{Vipula Rawte}
\author[4]{Ritvik Garimella}
\author[13]{Amitava Das}
\author[4]{Amit Sheth}
\author[11]{Vasu Sharma}
\author[12]{Aishwarya Naresh Reganti}
\author[11]{Vinija Jain}
\author[12]{Aman Chadha}
\address{$^1$Kalyani Government Engineering College, India. $^2$IIIT Delhi, India. $^3$BITS Pilani Hyderabad Campus, India. $^4$AI Institute, University of South Carolina, USA. $^5$IIIT Guwahati, India. $^6$NIT Silchar, India. $^7$San José State University, USA. $^8$UCLA, USA. $^9$Washington State University, USA. $^{10}$Vishwakarma Institute of Information Technology, India. $^{11}$Meta AI, USA. $^{12}$Amazon AI, USA. $^{13}$BITS Pilani Goa, India.}


\begin{abstract}
The growing capability of large language models to produce fluent, contextually coherent text has created mounting pressure on the systems and institutions responsible for ensuring the authenticity of digital content. Advanced generative models such as GPT-4, Claude 3.5, and Llama can produce highly coherent and human-like text, making it increasingly difficult to differentiate between human-written and AI-generated content. While these models have transformative applications, their misuse has raised concerns about misinformation, biased narratives, and security threats.

This paper provides a comprehensive analysis of state-of-the-art AI-generated text detection techniques and evaluates their effectiveness through the Counter Turing Test (CT2) shared tasks. Task A (Binary Classification) required participants to distinguish between human-written and AI-generated text, while Task B (Model Attribution) focused on identifying the specific language model responsible for generating a given text. The results demonstrated high performance in binary classification, with the top system achieving an F1 score of 1.0000, but significantly lower scores in model attribution, where the best system achieved 0.9531, highlighting the increased complexity of this task.

The top-performing teams leveraged fine-tuned transformer models, ensemble learning, and hybrid detection approaches, with DeBERTa-based and BART-based methods demonstrating strong results. However, the lower scores in Task B underscore the challenges of distinguishing outputs from different LLMs, necessitating further research into adversarial robustness, feature extraction, and cross-domain generalization.

\end{abstract}

\begin{keywords}
AI-Generated Text \sep Detection Techniques \sep Generative AI \sep Natural Language Processing
\end{keywords}

\maketitle

\input{introduction}
\input{related_work}
\input{shared_task}
\input{participating_systems}
\input{results}
\input{conclusion}

\bibliography{references}  

\end{document}

%% file: introduction.tex
\section{Introduction}
Generative AI technologies such as GPT-4 \cite{openai2023gpt4}, Claude \cite{anthropic2024claude}, and Llama \cite{touvron2023llama} have revolutionized the creation of synthetic text content. These tools leverage advanced neural architectures to produce highly coherent and contextually relevant text, enabling a wide range of applications in industries such as content creation, education, and customer service. However, the increasing accessibility of these tools has introduced significant challenges related to their potential misuse for spreading disinformation, generating spam, and manipulating public opinion \cite{solaiman2019release, krishna2023deception}.


High-profile incidents have demonstrated the societal and economic impact of AI-generated text. For instance, fabricated news articles and social media posts created by language models have manipulated political narratives, propagated false information, and influenced public sentiment. The first high-profile deployment of AI-generated audio in a U.S. electoral context occurred in January 2024, when a deepfake robocall impersonating President Biden's voice was distributed to thousands of New Hampshire voters ahead of the state primary, directing them not to vote \cite{nhdoj2024robocall}. Federal regulators subsequently proposed a \$6 million fine against the political consultant responsible \cite{fcc2024kramer}, demonstrating that AI-driven disinformation campaigns have transitioned from theoretical concern to documented electoral threat. 


The exponential growth in the capabilities of language models further exacerbates the problem. Technologies like GPT-4 and Claude have pushed the boundaries of linguistic fluency and contextual understanding, making it increasingly difficult to distinguish between AI-generated and human-written text using traditional detection methods. The proliferation of synthetic text has raised concerns among policymakers and technologists alike. A 2021 audit by the European Court of Auditors found that the EU Code of Practice on Disinformation fell short of holding online platforms accountable, noting that platforms' responsiveness to notifications from fact-checkers and researchers remained insufficient \cite{eca2021disinformation}.

Moreover, the misuse of AI-generated text extends beyond disinformation. It includes hate speech, phishing attacks, and the generation of biased or harmful narratives. For example, language models can be fine-tuned or prompted to produce content that embeds subtle biases or promotes divisive ideologies. These scenarios underscore the urgency of developing robust detection mechanisms to address the misuse of generative AI technologies.

This paper focuses on advancing detection techniques for AI-generated text by analyzing the state-of-the-art methods, identifying gaps, and proposing a comprehensive framework for evaluation. Building on the insights from the Defactify workshop series \cite{roy-2025-defactify-overview-text}, which has established itself as a leading forum for addressing challenges in multimodal fact-checking, this study aims to bridge the gap between academic research and practical applications. By consolidating findings across participating systems, we aim to contribute to the innovation of reliable, scalable detection systems that safeguard the integrity of digital platforms.

%% file: related_work.tex
\section{Related Work}

The evolution of Large Language Models (LLMs) has necessitated robust methods for distinguishing between human and AI-generated content. This section reviews existing datasets and detection methods.

\textbf{Statistical \& Zero-Shot Detectors:\\}
 In recent years, researchers are exploring methods to identify intrinsic signatures of AI text without requiring labeled training data. DetectGPT \cite{mitchell2023detectgptzeroshotmachinegeneratedtext} utilizes the negative curvature of an LLM’s log-probability surface to identify machine-generated text through probability discrepancies. Similarly, GLTR \cite{gehrmann2019gltrstatisticaldetectionvisualization} offers a suite of statistical tools, such as word-rank and entropy analysis, to improve human detection rates. More recent advancements include DNA-GPT \cite{yang2023dnagptdivergentngramanalysis}, which analyzes N-gram divergence during text "regeneration," and Binoculars \cite{hans2024spottingllmsbinocularszeroshot}, which calculates a score based on the contrast between two related language models to identify synthetic signals.

\textbf{Supervised Classifiers and Architectural Evolution: \\} This approach involves training neural discriminators on curated data of human and AI-generated text. Early work, such as Grover \cite{zellers2020defendingneuralfakenews}, demonstrated that the best generators often serve as the most effective discriminators for their own outputs. Research by Ippolito et al. \cite{ippolito2020automaticdetectiongeneratedtext} explored how various decoding strategies (e.g., nucleus sampling) create machine-detectable cues (or statistical artifacts) even when they successfully fool human evaluators. Ghostbuster \cite{verma2024ghostbusterdetectingtextghostwritten} utilizes a linear classifier trained on features extracted from weaker language models to perform black-box detection. Other studies have also utilized linguistic features (e.g., Linguistic Inquiry and Word Count) for authorship attribution \cite{uchendu-etal-2020-authorship}.

\textbf{Adversarial Evasion and Active Defense: \\} As detectors evolve, so do methods to bypass them. : Raidar \cite{mao2024raidargenerativeaidetection} identifies machine-generated text by measuring the character-level edit distance when an LLM rewrites a candidate passage, as AI-authored text tends to survive rewriting largely unchanged while human-written text undergoes substantially more revision. While paraphrasing remains a significant threat to detector accuracy, research suggests that semantic retrieval and caching API outputs can act as effective defenses \cite{krishna2023paraphrasingevadesdetectorsaigenerated}. To provide more definitive verification, watermarking frameworks, such as the "red list" and "green list" logit biasing proposed by Kirchenbauer et al. \cite{kirchenbauer2024watermarklargelanguagemodels}, embed invisible signals into LLM outputs. Conversely, frameworks like PIFE \cite{teja2025modelingattackdetectingaigenerated} aim for perturbation-invariant feature engineering to maintain detection accuracy despite character-level or word-level attacks.


%% file: shared_task.tex
\section{Task Details}

We use the dataset provided in \cite{roy2025comprehensivedatasethumanvs} for AI-generated text detection. The dataset included 73,193 samples across various domains, ensuring diversity in style, topic, and complexity.

\subsection{Data}

The dataset used in this shared task is fully documented in our companion dataset paper \cite{roy2025comprehensivedatasethumanvs}. It consists of 73,193 text samples drawn from authentic New York Times articles paired with synthetic counterparts generated by six state-of-the-art language models: Gemma-2-9b \cite{gemmateam2024gemma2}, Mistral-7B \cite{jiang2023mistral}, Qwen-2-72B \cite{yang2024qwen2}, LLaMA-3-8B \cite{grattafiori2024llama3}, Yi-Large \cite{young2024yi}, and GPT-4o \cite{openai2024gpt4o}. Each entry is accompanied by annotated metadata recording the source model, generation prompt, and relevant linguistic features, enabling both binary and multiclass evaluation. The data are partitioned into training, validation, and test sets of 51,247 / 10,983 / 10,963 samples respectively. We refer readers to \cite{roy2025comprehensivedatasethumanvs} for full dataset construction methodology, quality-control procedures, and exploratory statistics.


\subsection{Tasks}

\begin{itemize}
    \item \textbf{Task A: Binary Classification} \newline
    Participants were tasked with determining whether a given text sample was generated by AI or written by a human.  

    \item \textbf{Task B: Model Attribution} \newline
    Building on Task A, this task required participants to identify the specific language model responsible for generating a given text sample. Participants were provided with AI-generated samples and tasked with predicting which LLM generated the text. 

\end{itemize}

\subsection{Evaluation}

Performance in the competition is assessed using the \textbf{F$_1$-score}. 
For \textbf{Task A}, we report the \textbf{weighted F$_1$-score}, which accounts for label imbalance by averaging the F$_1$-scores of each class weighted by their support. 
For \textbf{Task B}, we use the \textbf{macro F$_1$-score}, which treats all classes equally by computing the unweighted mean of the per-class F$_1$-scores, thus emphasizing the ability to distinguish unique patterns across different model-generated outputs.







\subsection{Baseline}

We followed the baseline established in our companion dataset paper \cite{roy-2025-defactify-dataset-text}. We adopted a rewriting-based detection strategy derived from the Raidar framework \cite{mao2024raidargenerativeaidetection}. The underlying observation is that a language model asked to paraphrase a passage it originally generated will produce a version that is lexically close to the input, whereas the same model will make substantially more edits when paraphrasing text written by a human author. 
We prompted a fixed rewriting model (GPT-3.5-Turbo) to produce a meaning-preserving paraphrase of each input, then quantify the divergence between the original and the paraphrase using Levenshtein edit distance. For Task B (model attribution), we apply this procedure using each of the six candidate LLMs as the rewriter and assign the input to whichever model produces the paraphrase with the smallest edit distance. For Task A (binary classification), we classify the input as human-written when all six edit distances exceed a threshold; following \cite{roy2025comprehensivedatasethumanvs}, this threshold is set to the median of the per-model maximum edit distances observed across the training set. Table~\ref{tab:baseline_thresholds} reports baseline F1-scores under this strategy and two alternative threshold choices.



\begin{table}[h]
\centering
\begin{tabular}{|l|c|c|}
\hline
\textbf{Threshold Strategy} & \textbf{Task A (weighted F$_1$)} & \textbf{Task B (macro F$_1$)} \\
\hline
F$_1$-optimized       & 0.8400 & 0.0863 \\
Max edit distance      & 0.8457 & 0.0872 \\
Median edit distance   & 0.5300 & 0.0504 \\
\hline
\end{tabular}
\caption{Baseline F$_1$-scores under different threshold strategies for the Raidar rewriting-based method.}
\label{tab:baseline_thresholds}
\end{table}

\begin{center}
    \includegraphics[width=0.9\textwidth]{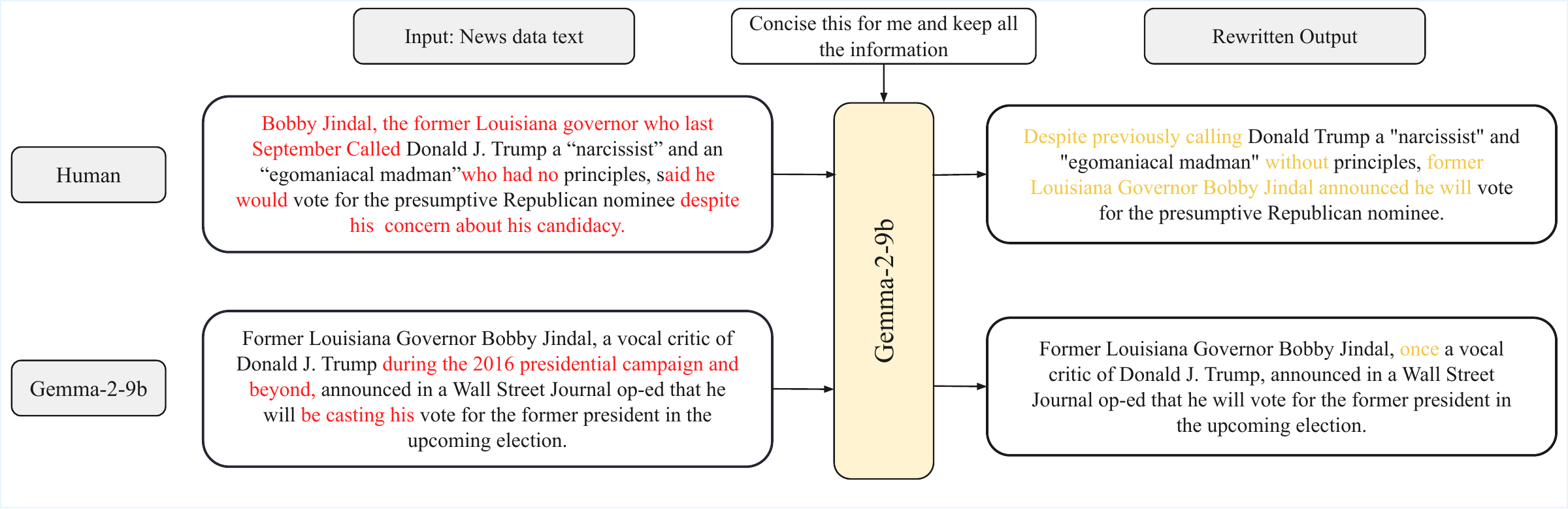}
    \captionof{figure}{\textbf{Illustration of Raidar concept.} Given a News data text and an LLM-generated text, the same LLM is asked to rewrite the inputs while preserving meaning. The rewriting of a human-written text undergoes more character-level edits (highlighted in red/yellow), while the rewriting of an LLM-generated text remains largely unchanged.}
    \label{fig:baseline_pipeline}
\end{center}

%% file: participating_systems.tex
\section{Participating Systems}
With over 52 registrations on the competition web page, there were final leaderboard submissions from 11 teams, with 7 teams making paper submissions.

The first participating team is \textbf{Sarang} \cite{trivedi2025sarang}. They present a fine-tuned DeBERTa-based \cite{he2021debertadecodingenhancedbertdisentangled} approach that secured first place in both Task A and Task B. Their method involves an ensemble of DeBERTa models trained on a noisy dataset, incorporating data augmentation techniques to enhance model robustness and generalization.

The \textbf{Dakiet} \cite{duong2025scalableframeworkclassifyingaigenerated} team introduces a scalable framework that integrates perceptual hashing, similarity measurement, and pseudo-labeling. Their approach leverages BART \cite{lewis2019bartdenoisingsequencetosequencepretraining} Large as the backbone model, achieving second place in Task A and third place in Task B.

The \textbf{Tesla} \cite{indurthi_identifying_machine_generated_text} team employs a feature-rich methodology, extracting style, language complexity, bias, subjectivity, and emotion-based features, alongside TF-IDF unigram and bigram representations. They utilize XGBoost \cite{Chen_2016} models, leading to high F1 scores and securing third place in Task A and second place in Task B.

The \textbf{SKDU} \cite{malviya2025skdu} team explores a pipelined approach leveraging RAIDAR-inspired rewriting features and NELA toolkit content-based features for feature extraction. Their experiments highlight that NELA features outperform RAIDAR\cite{mao2024raidargenerativeaidetection} features, with XGBoost proving to be the most effective classifier.

The \textbf{Drocks} \cite{abburi2025ai} team develops two neural architectures per task: an Optimized Model and a Simpler Variant. Their optimized model ranks 5th in Task A, while the simpler version secures 5th in Task B. The approach enhances a generalizable neural model with RoBERTa, BiLSTM, and E5 embeddings \cite{wang2024textembeddingsweaklysupervisedcontrastive} (Full Architecture). To reduce complexity, the Optimized Architecture replaces BiLSTM token-level features with stylometry. For multiclass classification, the Simple Architecture combines E5 embeddings and 11 stylometric features with a gradient boosting classifier.

The \textbf{AI\_Blues} \cite{guggilla2025ai} team adopts a fine-tuning–based strategy using both large language models and transformer encoders. They fine-tune GPT-4o-mini, LLaMA-3 8B \cite{grattafiori2024llama3herdmodels}, and BERT \cite{devlin2019bertpretrainingdeepbidirectional} for Task A and Task B, employing task-specific prompting for LLMs and supervised training for BERT. Their results demonstrate strong performance in human–AI discrimination, with GPT-4o-mini excelling in Task A, while BERT shows better performance in Task B.

The \textbf{Osint} \cite{agrahari2025tracing} team proposes \textbf{COT\_Finetuned}, a dual-task framework that integrates Chain-of-Thought (CoT) \cite{wei2023chainofthoughtpromptingelicitsreasoning} reasoning into supervised text classification. The approach jointly addresses AI-generated text detection and LLM identification, while producing interpretable reasoning traces. By incorporating CoT into models such as BERT, their method improves performance over standard fine-tuning and emphasizes explainability alongside accuracy.

%% file: results.tex
\section{Results}

\begin{table}[h]
    \centering
    \begin{tabular}{ccc}
        \toprule
        \textbf{S.No} & \textbf{Team Name} & \textbf{Scores} \\
        \midrule
        1  & \textbf{Sarang}        & \textbf{1.0000}   \\
        2  & Dakiet        & 0.9999   \\
        3  & Tesla         & 0.9962   \\
        4  & SKDU          & 0.9945   \\
        5  & Drocks        & 0.9941   \\
        6  & Llama\_Mamba   & 0.9880   \\
        7  & AI\_Blues     & 0.9547   \\
        8  & NLP\_great    & 0.9157   \\
        9  & Osint         & 0.8982   \\
        10 & Xiaoyu        & 0.8030   \\
        11 & Rohan         & 0.7546   \\
         - & \textbf{BASELINE} & \textbf{0.5300}  \\
        \bottomrule
    \end{tabular}
    \caption{Leaderboard for Task A: Classify each text document as either AI-generated or human-written.}
    \label{tab:task_a}
\end{table}

Table \ref{tab:task_a} showcases the leaderboard for Task A, where participants are ranked based on their scores. The highest score of 1.0000 is achieved by Team Sarang, followed closely by Team Dakiet with a score of 0.9999. The top five participants have scores above 0.99, demonstrating strong performance in this task.

\begin{table}[h]
    \centering
    \begin{tabular}{cccc}
        \toprule
        \textbf{S.No}  & \textbf{Team Name} & \textbf{Scores} \\
        \midrule
        1  & \textbf{Sarang}        & \textbf{0.9531}   \\
        2  & Tesla         & 0.9218   \\
        3  & Dakiet        & 0.9082   \\
        4  & SKDU          & 0.7615   \\
        5  & Drocks        & 0.6270   \\
        6  & Xiaoyu        & 0.5696   \\
        7  & AI\_Blues     & 0.4698   \\
        8  & Llama\_Mamba   & 0.4551   \\
        9  & Rohan         & 0.4053   \\
        10 & Osint         & 0.3072   \\
        11 & NLP\_great    & 0.1874   \\
         -   & \textbf{BASELINE}    & \textbf{0.0504}  \\
        \bottomrule
    \end{tabular}
    \caption{Leaderboard for Task B: Given an AI-generated text, determine which specific LLM produced it. }
    \label{tab:task_b}
\end{table}

\FloatBarrier 

Table \ref{tab:task_b} presents the leaderboard for Task B, which has a different ranking compared to Task A. Team Sarang secures the top position with a score of 0.9531. The overall scores for Task B are lower than those in Task A, confirming that Task B is more challenging. The top three participants get comparable scores whereas the lower-ranked participants have considerably lower scores.

%% file: conclusion.tex
\section{Conclusion}

The Defactify 4.0 workshop has highlighted the urgent need for advanced methodologies to detect AI-generated text. The shared tasks demonstrated that while binary classification of AI-generated text has seen strong performance, model attribution remains a significant challenge, with lower accuracy across all participants. The top-performing systems leveraged fine-tuned transformer models, ensemble learning, and hybrid techniques, underscoring the importance of combining linguistic and feature-based detection methods. By building on the shared tasks, datasets, and evaluation frameworks discussed in this paper, we aim to inspire further innovation in this critical area. Future research should focus on improving detection robustness, enhancing adversarial resistance, and developing scalable solutions for real-world deployment. As generative models continue to improve, building detection systems that are accurate, adversarially robust, and domain-generalizable will only become more critical. This shared task takes a concrete step toward that objective, surfacing what current methods can reliably achieve and where meaningful gaps remain.

%% file: references.bib
@misc{mao2024raidargenerativeaidetection,
      title={Raidar: geneRative AI Detection viA Rewriting}, 
      author={Chengzhi Mao and Carl Vondrick and Hao Wang and Junfeng Yang},
      year={2024},
      eprint={2401.12970},
      archivePrefix={arXiv},
      primaryClass={cs.CL},
      url={https://arxiv.org/abs/2401.12970}, 
}

@misc{anthropic2024claude,
  author    = {Anthropic},
  title     = {Claude AI: Conversational AI Assistant},
  year      = {2024},
  howpublished = {\url{https://www.anthropic.com/claude}},
  note      = {Accessed: 2025-01-25}
}

@article{solaiman2019release,
  author    = {Solaiman, Irene and Brundage, Miles},
  title     = {Release Strategies and the Social Impacts of Language Models},
  journal   = {OpenAI Technical Report},
  year      = {2019}
}

@article{krishna2023deception,
  author    = {Krishna, Prakhar and others},
  title     = {Deception in AI-Generated Text: Adversarial Evaluation},
  journal   = {ACL Workshop on Fact-Checking},
  year      = {2023}
}

@inproceedings{roy-2025-defactify-overview-text,  title={Overview of Text Counter Turing Test: AI Generated Text Detection},  author={Rajarshi Roy and Gurpreet Singh and Ashhar Aziz and Shashwat Bajpai and Nasrin Imanpour and Shwetangshu Biswas and Kapil Wanaskar and  Parth Patwa and Subhankar Ghosh and Shreyas Dixit and Nilesh Ranjan Pal and Vipula Rawte and Ritvik Garimella and Amitava Das and Amit Sheth and Vasu Sharma and Aishwarya Naresh Reganti and Vinija Jain and Aman Chadha},  year={2025},   publisher = {CEUR},  booktitle = {proceedings of {D}e{F}actify 4: Fourth workshop on Multimodal Fact-Checking and Hate Speech Detection}}

@inproceedings{roy-2025-defactify-dataset-text,  title={Defactify-Text: A Comprehensive Dataset for Human vs. AI Generated Text Detection},  author={Rajarshi Roy and Gurpreet Singh  and Ashhar Aziz and Shashwat Bajpai and Nasrin Imanpour and Shwetangshu Biswas and Kapil Wanaskar and Parth Patwa and Subhankar Ghosh and Shreyas Dixit and Nilesh Ranjan Pal and Vipula Rawte and Ritvik Garimella and Amitava Das and Amit Sheth and Vasu Sharma and Aishwarya Naresh Reganti and Vinija Jain and Aman Chadha},  year={2025},   publisher = {CEUR},  booktitle = {proceedings of {D}e{F}actify 4: Fourth workshop on Multimodal Fact-Checking and Hate Speech Detection}}

@misc{roy2025comprehensivedatasethumanvs,
      title={A Comprehensive Dataset for Human vs. AI Generated Text Detection}, 
      author={Rajarshi Roy and Gurpreet Singh and Ashhar Aziz and Shashwat Bajpai and Nasrin Imanpour and Shwetangshu Biswas and Kapil Wanaskar and Parth Patwa and Subhankar Ghosh and Shreyas Dixit and Nilesh Ranjan Pal and Vipula Rawte and Ritvik Garimella and Gaytri Jena and Amitava Das and Amit Sheth and Vasu Sharma and Aishwarya Naresh Reganti and Vinija Jain and Aman Chadha},
      year={2025},
      eprint={2510.22874},
      archivePrefix={arXiv},
      primaryClass={cs.CL},
      url={https://arxiv.org/abs/2510.22874}, 
}

@article{touvron2023llama,
  author    = {Hugo Touvron and others},
  title     = {LLaMA: Open and Efficient Foundation Language Models},
  journal   = {arXiv preprint arXiv:2302.13971},
  year      = {2023},
  url       = {https://arxiv.org/abs/2302.13971}
}

@article{openai2023gpt4,
  author    = {OpenAI},
  title     = {GPT-4 Technical Report},
  journal   = {OpenAI Technical Report},
  year      = {2023},
  url       = {https://arxiv.org/abs/2303.08774}
}

@article{grattafiori2024llama3herdmodels,
  title={The llama 3 herd of models},
  author={Grattafiori, Aaron and Dubey, Abhimanyu and Jauhri, Abhinav and Pandey, Abhinav and Kadian, Abhishek and Al-Dahle, Ahmad and Letman, Aiesha and Mathur, Akhil and Schelten, Alan and Vaughan, Alex and others},
  journal={arXiv preprint arXiv:2407.21783},
  year={2024}
}

@misc{he2021debertadecodingenhancedbertdisentangled,
      title={DeBERTa: Decoding-enhanced BERT with Disentangled Attention}, 
      author={Pengcheng He and Xiaodong Liu and Jianfeng Gao and Weizhu Chen},
      year={2021},
      eprint={2006.03654},
      archivePrefix={arXiv},
      primaryClass={cs.CL},
      url={https://arxiv.org/abs/2006.03654}, 
}

@misc{lewis2019bartdenoisingsequencetosequencepretraining,
      title={BART: Denoising Sequence-to-Sequence Pre-training for Natural Language Generation, Translation, and Comprehension}, 
      author={Mike Lewis and Yinhan Liu and Naman Goyal and Marjan Ghazvininejad and Abdelrahman Mohamed and Omer Levy and Ves Stoyanov and Luke Zettlemoyer},
      year={2019},
      eprint={1910.13461},
      archivePrefix={arXiv},
      primaryClass={cs.CL},
      url={https://arxiv.org/abs/1910.13461}, 
}

@inproceedings{Chen_2016, series={KDD ’16},
   title={XGBoost: A Scalable Tree Boosting System},
   url={http://dx.doi.org/10.1145/2939672.2939785},
   DOI={10.1145/2939672.2939785},
   booktitle={Proceedings of the 22nd ACM SIGKDD International Conference on Knowledge Discovery and Data Mining},
   publisher={ACM},
   author={Chen, Tianqi and Guestrin, Carlos},
   year={2016},
   month=aug, pages={785–794},
   collection={KDD ’16} }

@misc{wang2024textembeddingsweaklysupervisedcontrastive,
      title={Text Embeddings by Weakly-Supervised Contrastive Pre-training}, 
      author={Liang Wang and Nan Yang and Xiaolong Huang and Binxing Jiao and Linjun Yang and Daxin Jiang and Rangan Majumder and Furu Wei},
      year={2024},
      eprint={2212.03533},
      archivePrefix={arXiv},
      primaryClass={cs.CL},
      url={https://arxiv.org/abs/2212.03533}, 
}

@misc{devlin2019bertpretrainingdeepbidirectional,
      title={BERT: Pre-training of Deep Bidirectional Transformers for Language Understanding}, 
      author={Jacob Devlin and Ming-Wei Chang and Kenton Lee and Kristina Toutanova},
      year={2019},
      eprint={1810.04805},
      archivePrefix={arXiv},
      primaryClass={cs.CL},
      url={https://arxiv.org/abs/1810.04805}, 
}

@misc{wei2023chainofthoughtpromptingelicitsreasoning,
      title={Chain-of-Thought Prompting Elicits Reasoning in Large Language Models}, 
      author={Jason Wei and Xuezhi Wang and Dale Schuurmans and Maarten Bosma and Brian Ichter and Fei Xia and Ed Chi and Quoc Le and Denny Zhou},
      year={2023},
      eprint={2201.11903},
      archivePrefix={arXiv},
      primaryClass={cs.CL},
      url={https://arxiv.org/abs/2201.11903}, 
}

@article{agrahari2025tracing,
  title={Tracing Thought: Using Chain-of-Thought Reasoning to Identify the LLM Behind AI-Generated Text},
  author={Agrahari, Shifali and Singh, Sanasam Ranbir},
  journal={arXiv preprint arXiv:2504.16913},
  year={2025}
}

@article{trivedi2025sarang,
  title={Sarang at DEFACTIFY 4.0: Detecting AI-Generated Text Using Noised Data and an Ensemble of DeBERTa Models},
  author={Trivedi, Avinash and Sivanesan, Sangeetha},
  journal={arXiv preprint arXiv:2502.16857},
  year={2025}
}

@article{malviya2025skdu,
  title={Skdu at de-factify 4.0: Natural language features for ai-generated text-detection},
  author={Malviya, Shrikant and Arnau-Gonz{\'a}lez, Pablo and Arevalillo-Herr{\'a}ez, Miguel and Katsigiannis, Stamos},
  journal={arXiv preprint arXiv:2503.22338},
  year={2025}
}

@article{guggilla2025ai,
  title={AI Generated Text Detection Using Instruction Fine-tuned Large Language and Transformer-Based Models},
  author={Guggilla, Chinnappa and Roy, Budhaditya and Chavan, Trupti Ramdas and Rahman, Abdul and Bowen, Edward},
  journal={arXiv preprint arXiv:2507.05157},
  year={2025}
}

@article{abburi2025ai,
  title={AI-generated text detection: A multifaceted approach to binary and multiclass classification},
  author={Abburi, Harika and Bhattacharya, Sanmitra and Bowen, Edward and Pudota, Nirmala},
  journal={arXiv preprint arXiv:2505.11550},
  year={2025}
}

@misc{duong2025scalableframeworkclassifyingaigenerated,
      title={Scalable Framework for Classifying AI-Generated Content Across Modalities}, 
      author={Anh-Kiet Duong and Petra Gomez-Krämer},
      year={2025},
      eprint={2502.00375},
      archivePrefix={arXiv},
      primaryClass={cs.CV},
      url={https://arxiv.org/abs/2502.00375}, 
}

@misc{mitchell2023detectgptzeroshotmachinegeneratedtext,
      title={DetectGPT: Zero-Shot Machine-Generated Text Detection using Probability Curvature}, 
      author={Eric Mitchell and Yoonho Lee and Alexander Khazatsky and Christopher D. Manning and Chelsea Finn},
      year={2023},
      eprint={2301.11305},
      archivePrefix={arXiv},
      primaryClass={cs.CL},
      url={https://arxiv.org/abs/2301.11305}, 
}

@misc{gehrmann2019gltrstatisticaldetectionvisualization,
      title={GLTR: Statistical Detection and Visualization of Generated Text}, 
      author={Sebastian Gehrmann and Hendrik Strobelt and Alexander M. Rush},
      year={2019},
      eprint={1906.04043},
      archivePrefix={arXiv},
      primaryClass={cs.CL},
      url={https://arxiv.org/abs/1906.04043}, 
}

@misc{hans2024spottingllmsbinocularszeroshot,
      title={Spotting LLMs With Binoculars: Zero-Shot Detection of Machine-Generated Text}, 
      author={Abhimanyu Hans and Avi Schwarzschild and Valeriia Cherepanova and Hamid Kazemi and Aniruddha Saha and Micah Goldblum and Jonas Geiping and Tom Goldstein},
      year={2024},
      eprint={2401.12070},
      archivePrefix={arXiv},
      primaryClass={cs.CL},
      url={https://arxiv.org/abs/2401.12070}, 
}

@inproceedings{uchendu-etal-2020-authorship,
    title = "Authorship Attribution for Neural Text Generation",
    author = "Uchendu, Adaku  and
      Le, Thai  and
      Shu, Kai  and
      Lee, Dongwon",
    editor = "Webber, Bonnie  and
      Cohn, Trevor  and
      He, Yulan  and
      Liu, Yang",
    booktitle = "Proceedings of the 2020 Conference on Empirical Methods in Natural Language Processing (EMNLP)",
    month = nov,
    year = "2020",
    address = "Online",
    publisher = "Association for Computational Linguistics",
    url = "https://aclanthology.org/2020.emnlp-main.673/",
    doi = "10.18653/v1/2020.emnlp-main.673",
    pages = "8384--8395",
   }

@misc{verma2024ghostbusterdetectingtextghostwritten,
      title={Ghostbuster: Detecting Text Ghostwritten by Large Language Models}, 
      author={Vivek Verma and Eve Fleisig and Nicholas Tomlin and Dan Klein},
      year={2024},
      eprint={2305.15047},
      archivePrefix={arXiv},
      primaryClass={cs.CL},
      url={https://arxiv.org/abs/2305.15047}, 
}

@misc{ippolito2020automaticdetectiongeneratedtext,
      title={Automatic Detection of Generated Text is Easiest when Humans are Fooled}, 
      author={Daphne Ippolito and Daniel Duckworth and Chris Callison-Burch and Douglas Eck},
      year={2020},
      eprint={1911.00650},
      archivePrefix={arXiv},
      primaryClass={cs.CL},
      url={https://arxiv.org/abs/1911.00650}, 
}

@misc{krishna2023paraphrasingevadesdetectorsaigenerated,
      title={Paraphrasing evades detectors of AI-generated text, but retrieval is an effective defense}, 
      author={Kalpesh Krishna and Yixiao Song and Marzena Karpinska and John Wieting and Mohit Iyyer},
      year={2023},
      eprint={2303.13408},
      archivePrefix={arXiv},
      primaryClass={cs.CL},
      url={https://arxiv.org/abs/2303.13408}, 
}

@misc{kirchenbauer2024watermarklargelanguagemodels,
      title={A Watermark for Large Language Models}, 
      author={John Kirchenbauer and Jonas Geiping and Yuxin Wen and Jonathan Katz and Ian Miers and Tom Goldstein},
      year={2024},
      eprint={2301.10226},
      archivePrefix={arXiv},
      primaryClass={cs.LG},
      url={https://arxiv.org/abs/2301.10226}, 
}

@misc{teja2025modelingattackdetectingaigenerated,
      title={Modeling the Attack: Detecting AI-Generated Text by Quantifying Adversarial Perturbations}, 
      author={Lekkala Sai Teja and Annepaka Yadagiri and Sangam Sai Anish and Siva Gopala Krishna Nuthakki and Partha Pakray},
      year={2025},
      eprint={2510.02319},
      archivePrefix={arXiv},
      primaryClass={cs.CR},
      url={https://arxiv.org/abs/2510.02319}, 
}

@misc{zellers2020defendingneuralfakenews,
      title={Defending Against Neural Fake News}, 
      author={Rowan Zellers and Ari Holtzman and Hannah Rashkin and Yonatan Bisk and Ali Farhadi and Franziska Roesner and Yejin Choi},
      year={2020},
      eprint={1905.12616},
      primaryClass={cs.CL},
      url={https://arxiv.org/abs/1905.12616}, 
}

@misc{indurthi_identifying_machine_generated_text,
  title        = {Identifying Machine Generated Text with Stylometric Features},
  author       = {Vijayasaradhi Indurthi and Vasudeva Varma},
  institution  = {International Institute of Information Technology, Hyderabad},
  note = {proceedings of {D}e{F}actify 4: Fourth workshop on Multimodal Fact-Checking and Hate Speech Detection},
  journal={proceedings of DeFactify},
  volume={4}
}

@misc{nhdoj2024robocall,
  author       = {{New Hampshire Department of Justice, Office of the Attorney General}},
  title        = {Voter Suppression {AI} Robocall Investigation Update},
  howpublished = {Press release},
  month        = feb,
  year         = {2024},
  url          = {https://www.doj.nh.gov/news-and-media/voter-suppression-ai-robocall-investigation-update}
}

@misc{fcc2024kramer,
  author       = {{Federal Communications Commission}},
  title        = {In the Matter of {Steve Kramer}: Notice of Apparent Liability for Forfeiture},
  howpublished = {{FCC} 24-59, File No.\ EB-TCD-24-00036094},
  month        = may,
  year         = {2024},
  url          = {https://docs.fcc.gov/public/attachments/FCC-24-59A1.pdf}
}

@misc{eca2021disinformation,
  author       = {{European Court of Auditors}},
  title        = {Disinformation Affecting the {EU}: Tackled but Not Tamed},
  howpublished = {Special Report No.\ 09/2021, Publications Office of the European Union},
  month        = jun,
  year         = {2021},
  doi          = {10.2865/337863},
  url          = {https://op.europa.eu/webpub/eca/special-reports/disinformation-9-2021/en/}
}

@article{gemmateam2024gemma2,
  author       = {{Gemma Team}},
  title        = {{Gemma 2}: Improving Open Language Models at a Practical Size},
  journal      = {arXiv preprint arXiv:2408.00118},
  year         = {2024},
  eprint       = {2408.00118},
  archivePrefix= {arXiv},
  primaryClass = {cs.CL},
  url          = {https://arxiv.org/abs/2408.00118}
}

@article{jiang2023mistral,
  author       = {Jiang, Albert Q. and Sablayrolles, Alexandre and Mensch, Arthur
                  and Bamford, Chris and Chaplot, Devendra Singh and de las Casas, Diego
                  and Bressand, Florian and Lengyel, Gianna and Lample, Guillaume
                  and Saulnier, Lucile and Lavaud, L{\'e}lio Renard and Lachaux, Marie-Anne
                  and Stock, Pierre and Scao, Teven Le and Lavril, Thibaut
                  and Wang, Thomas and Lacroix, Timoth{\'e}e and Sayed, William El},
  title        = {Mistral {7B}},
  journal      = {arXiv preprint arXiv:2310.06825},
  year         = {2023},
  eprint       = {2310.06825},
  archivePrefix= {arXiv},
  primaryClass = {cs.CL},
  doi          = {10.48550/arXiv.2310.06825},
  url          = {https://arxiv.org/abs/2310.06825}
}

@article{yang2024qwen2,
  author       = {Yang, An and Yang, Baosong and Hui, Binyuan and Zheng, Bo
                  and Yu, Bowen and Zhou, Chang and Li, Chengpeng and Li, Chengyuan
                  and Liu, Dayiheng and Huang, Fei and others},
  title        = {{Qwen2} Technical Report},
  journal      = {arXiv preprint arXiv:2407.10671},
  year         = {2024},
  eprint       = {2407.10671},
  archivePrefix= {arXiv},
  primaryClass = {cs.CL},
  doi          = {10.48550/arXiv.2407.10671},
  url          = {https://arxiv.org/abs/2407.10671}
}

@article{grattafiori2024llama3,
  author       = {Grattafiori, Aaron and Dubey, Abhimanyu and Jauhri, Abhinav
                  and Pandey, Abhinav and Kadian, Abhishek and Al-Dahle, Ahmad
                  and Letman, Aiesha and Mathur, Akhil and Schelten, Alan
                  and Vaughan, Alex and others},
  title        = {The {Llama 3} Herd of Models},
  journal      = {arXiv preprint arXiv:2407.21783},
  year         = {2024},
  eprint       = {2407.21783},
  archivePrefix= {arXiv},
  primaryClass = {cs.AI},
  doi          = {10.48550/arXiv.2407.21783},
  url          = {https://arxiv.org/abs/2407.21783}
}

@article{young2024yi,
  author       = {{01.AI} and Young, Alex and Chen, Bei and Li, Chao
                  and Huang, Chengen and Zhang, Ge and Zhang, Guanwei
                  and Li, Heng and Zhu, Jiangcheng and Chen, Jianqun
                  and Chang, Jing and Yu, Kaidong and Liu, Peng and Liu, Qiang
                  and Yue, Shawn and Yang, Senbin and Yang, Shiming and Yu, Tao
                  and Xie, Wen and Huang, Wenhao and Hu, Xiaohui and Ren, Xiaoyi
                  and Niu, Xinyao and Nie, Pengcheng and Xu, Yuchi and Liu, Yudong
                  and Wang, Yue and Cai, Yuxuan and Gu, Zhenyu
                  and Liu, Zhiyuan and Dai, Zonghong},
  title        = {{Yi}: Open Foundation Models by {01.AI}},
  journal      = {arXiv preprint arXiv:2403.04652},
  year         = {2024},
  eprint       = {2403.04652},
  archivePrefix= {arXiv},
  primaryClass = {cs.CL},
  url          = {https://arxiv.org/abs/2403.04652}
}

@misc{openai2024gpt4o,
  author       = {{OpenAI}},
  title        = {Hello {GPT-4o}},
  howpublished = {OpenAI Blog},
  month        = may,
  year         = {2024},
  url          = {https://openai.com/index/hello-gpt-4o/}
}

@misc{yang2023dnagptdivergentngramanalysis,
      title={DNA-GPT: Divergent N-Gram Analysis for Training-Free Detection of GPT-Generated Text}, 
      author={Xianjun Yang and Wei Cheng and Yue Wu and Linda Petzold and William Yang Wang and Haifeng Chen},
      year={2023},
      eprint={2305.17359},
      archivePrefix={arXiv},
      primaryClass={cs.CL},
      url={https://arxiv.org/abs/2305.17359}, 
}
